\pdfoutput=1

\documentclass[11pt]{article}

\usepackage{acl}

\usepackage{times}
\usepackage{latexsym}
\usepackage{graphicx}
\usepackage{times}

\usepackage[T1]{fontenc}

\usepackage[utf8]{inputenc}

\usepackage{microtype}
\usepackage{hyperref}

\title{Hate Speech Detection and Classification in Amharic Text \\ with Deep Learning}

\author{Samuel Minale Gashe \\
Department of ComputerSc. \\
Addis Ababa University \\
Addis Ababa, Ethiopia \\
\texttt{samuel.minale}\\
\texttt{@aau.edu.et}\\\And
Seid Muhie Yimam \\
Department of Informatics\\
Universität Hamburg \\
Hamburg, Germany\\
\texttt{seid.muhie.yimam} \\
\texttt{@uni-hamburg.de} \And
Yaregal Assabie \\
Department of Computer Sci. \\
Addis Ababa University\\
Addis Ababa, Ethiopia\\
\texttt{yaregal.assabi} \\
\texttt{@aau.edu.et} \\ }

\begin{document}
\maketitle {}
\begin{abstract}
Hate speech is a growing problem on social media. It can seriously impact society, especially in countries like Ethiopia, where it can trigger conflicts among diverse ethnic and religious groups. While hate speech detection in resource rich languages are progressing, for low resource languages such as Amharic are lacking. To address this gap, we develop Amharic hate speech data and SBi-LSTM deep learning model that can detect and classify  text into four categories of hate speech: racial, religious, gender, and non-hate speech. We have annotated 5k Amharic social media post and comment data into four categories. The data is annotated using a custom annotation tool by a total of 100 native Amharic speakers.  The model achieves a 94.8 F1-score performance. Future improvements will include expanding the dataset and develop state-of-the art models.


\end{abstract}

\section{Introduction}
Social media platforms such as Facebook allow anyone to share content quickly and easily. This has led to an increase in the amount of content that is shared. While this can be a positive contribution, it has also led to an increase in abusive and hateful speech particularly in Amharic tweets when compared to other Ethiopian languages \citep{1_yimam2019analysis}. This is because Amharic is a widely spoken and working language in Ethiopia that has been used as a lingua franca of Ethiopia \citep{3_boston_unv_lib} for centuries. It is a low-resource language with a complex morphology \citep{4_abate2014development} but it is becoming increasingly important as a medium for online communication.

Hate speech can have real-world psychological, physical, and offline impacts on victims. According to the Center for Advancement of Rights and Democracy \citep{2_card_book}, failing to detect and penalize hate speech speakers can have serious consequences. Hate speech can start as simple jokes or rumors but it can quickly escalate to violence and genocide. It is important to detect and control hate speech dissemination in its early stages.

Hate speech is a serious problem in Ethiopia, where it is often spread online unknowingly or as a form of free speech. The government has implemented a law against hate speech, but it is difficult to enforce without an automated detection mechanism.

Previous research works on Amharic hate speech \cite{1_yimam2019analysis, 5_mossie2018social} did not analyze different types of hate speech and not applied different types of normalization to their data. As the researchers reported this led their systems to misclassify text as hate speech.

To develop an effective hate speech detection system, we reviewed, created new dataset and developed three model types using our new dataset: rule-based, classical machine learning (ML), and deep learning techniques. We found that deep learning techniques, specifically SBi-LSTM \citep{6_do2019hate, 7_graves2013hybrid, 8_lecun2015deep} is the most effective for Amharic hate speech detection.

\section{Related Work}
Researchers have developed methods to detect hate speech in different languages. We have reviewed some languages including English \citep{9_badjatiya2017deep,10_zimmerman-etal-2018-improving}, Danish \citep{11_sigurbergsson2019offensive}, Dutch \citep{12_tulkens2016dictionary}, German \citep{13_jaki2019right}, Indonesian \citep{14_alfina2017hate}, Italian \citep{15_del2017hate}, Vietnamese \citep{16_luu2020comparison}, Arabic \citep{17_albadi2018they}, and Hebrew \citep{18_warner2012detecting}. 
As the above researchers reveal classic ML techniques achieved satisfactory results, but deep learning techniques achieved higher values. However, even there are few studies on Amharic hate speech detection, but their accuracy is low, the dataset is still not available, and their methods are not state-of-the-art \citep{1_yimam2019analysis, 5_mossie2018social, 19_yonas2019hate}.

\section{Design of the System}
\begin{figure}
    \centering
    \includegraphics[width=6cm, height=6cm]{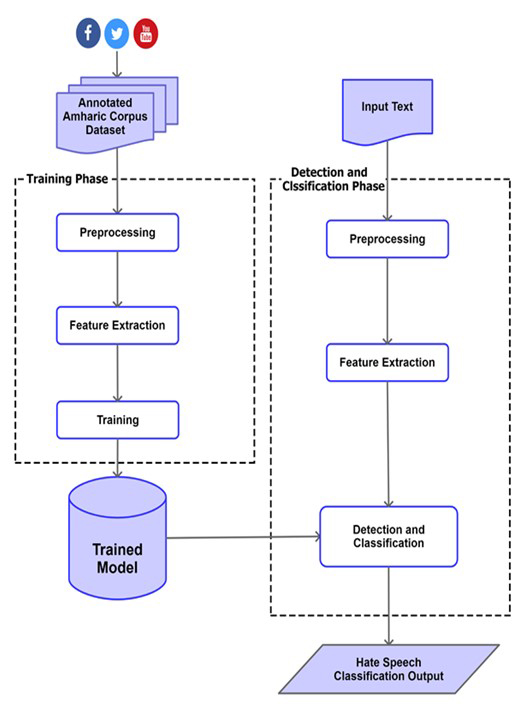}
    \caption{Proposed hate speech detection architecture}
    \label{fig:architectural_design}
\end{figure}

The proposed architecture accepts labeled Amharic hate speech text and classifies it into racial, religious, gender, and non-hate speech. The flow and the order in which those components are organized and form the system are clearly shown in Figure \ref{fig:architectural_design}.

The first component of the architecture is a pre-processing component that cleans and prepares the input dataset. The second component balances the number of texts using Synthetic Minority Oversampling Technique \citep{21_chawla2002smote} to increase the performance of the system. The third component does word representation and feature extraction using fastText, neural network tool, and TF-IDF embedding technique. The fourth component trains the model with an output of a knowledge base model which can be used independently, and the last component is responsible for classification of pre-processed and feature-extracted input text into the corresponding hate speech categories using the trained model knowledge base output. This component of the architecture incorporates rectified linear activation unit to identify complex relationships and Softmax activation functions for multi-class classifications.

\section{Dataset Collection and Preparation}

We collected above 1 million posts and comments from Twitter (using Twitter API), Facebook (using Facepager tool \citep{20_facepager}), and YouTube. The data is collected from Aug 2014 to June 2022. After the data was collected, we applied different pre-processing techniques such as cleaning, consolidating, and filtering Amharic text, and became ready for annotation. In the cleaning process, we applied PYCLD2 language identifier Python library. Finally, the dataset to be annotated was filtered to 5k using carefully selected keywords related to hate speech, offensive language, religion, and gender.

\section{Dataset Annotation}
After developing a clear annotation guideline, 100 native Amharic speaker annotators participated who have different demographic and socio-cultural backgrounds for manual annotation. For annotation, we developed a custom-built tool called \href{https://annotate.shegerapps.com/}{“Amharic Hate Speech Annotation Tool”}. This tool streamlines team annotation controls data preparation, and makes it easier for dataset import and export. 

\section{Experimental Setup}
To implement the proposed design of Figure \ref{fig:architectural_design}, different Python 3 frameworks are used like Keras, Scikit-learn, NumPy, TensorFlow, Pandas, and NLTK. After loading the annotated dataset, data cleaning, normalization, and tokenization are done in the pre-processing.

We then developed a deep learning model based on a Stacked Bidirectional Long Short-Term Memory (SBi-LSTM) architecture on Google Collab environment. We tuned the hyperparameters of the model to achieve F1-Score of 94.8. Our model outperforms our baseline rule-based methods resulting F1-Score of 40.1 and classic machine-learning approaches resulting F1-Score of 80.3.
\section{Conclusion and Recommendation}
In this paper, we have developed a deep learning system for detecting and classifying Amharic hate speech. Our system achieved an F1-Score of 94.8, which is a significant improvement over previous works. We believe that our system can be used to improve the detection of hate speech on social media in Ethiopia. Our new dataset, preprocessing scripts, and annotation tools will be available publicly upon the acceptance of the paper.

We recommend further developing our system to identify targeted ethnicities and religions, support more hate speech categories, integrate it with different social media platforms, and automate blocking of hate speech on different social media platforms.

\bibliographystyle{acl_natbib}
\bibliography{myref}

\end{document}